\documentclass[letterpaper, 10 pt, conference]{ieeeconf}  

\IEEEoverridecommandlockouts                              
\usepackage{amsmath}
\usepackage{amssymb}
\usepackage{color}
\usepackage{enumerate}
\usepackage{graphicx}
\usepackage{caption}
\usepackage{subcaption}
\usepackage{url}

\usepackage{fancyhdr}
\fancypagestyle{withfooter}{
  
  \fancyfoot[C]{\footnotesize Accepted to the IEEE IROS 2024 workshop on Long-Term Perception for Autonomy in Dynamic Human-shared Environments: What Do Robots Need?}
}

\title{\LARGE \bf
Moving Object Segmentation in Point Cloud Data using Hidden Markov Models
}

\author{Vedant Bhandari, Jasmin James, Tyson Phillips and P. Ross McAree%
    \thanks{The University of Queensland, 4072, Australia {\tt\small v.bhandari@uq.edu.au}}%
}

\begin{document}

\maketitle
\thispagestyle{withfooter}
\pagestyle{withfooter}

\begin{abstract}
Autonomous agents require the capability to identify dynamic objects in their environment for safe planning and navigation.
Incomplete and erroneous dynamic detections jeopardize the agent's ability to accomplish its task.
Dynamic detection is a challenging problem due to the numerous sources of uncertainty inherent in the problem's inputs and the wide variety of applications, which often lead to use-case-tailored solutions.
We propose a robust learning-free approach to segment moving objects in point cloud data.
The foundation of the approach lies in modelling each voxel using a hidden Markov model (HMM), and probabilistically integrating beliefs into a map using an HMM filter.
The proposed approach is tested on benchmark datasets and consistently performs better than or as well as state-of-the-art methods with strong generalized performance across sensor characteristics and environments.
The approach is open-sourced at \mbox{\url{https://github.com/vb44/HMM-MOS}}.
\end{abstract}

\vspace{-3mm}
\section{INTRODUCTION}
\label{sec:introduction}
Detecting motion in the workspace is a crucial capability for autonomous agents.
Agents employ sensors such as cameras and Light Detection and Ranging (LiDAR) to image their environment.
The Moving Object Segmentation (MOS) problem involves categorizing the pixels in an image or the points in a LiDAR scan as static or dynamic.
A key challenge is to provide consistent detection across environments, platform dynamics, and sensor characteristics.
There is a need for a solution that offers generalized and accurate dynamic detection.
To address this, we propose a learning-free MOS approach demonstrating strong generalized performance.

\vspace{-1mm}
\section{RELATED WORK}
\label{sec:related_work}
Learning-free approaches to solving the MOS problem are generally categorized as scan-based or map-based.

Scan-based methods compare successive observations to highlight discrepancies in the environment.
Underwood et al.~\cite{Underwood2013} detect changes in scans by identifying discrepancies in the observed space, with points labelled dynamic if they are greater than a distance from previously registered points.
Yoon et al.~\cite{Yoon2019} use a similar idea with dynamic detection relying on a window size that allows sufficient displacement of the object - a characteristic differing between object classes.
Mersch et al.~\cite{Mersch2022} demonstrate state-of-the-art performance with 4DMOS using sparse 4D spatio-temporal convolutions for segmenting dynamic points in scan-to-scan comparisons.

Map-based methods construct a representation of the environment and query changes in occupancy.
Octomap by Armin et al.~\cite{Hornung2013} clamp the occupancy probabilities to evolve beliefs in dynamic environments.
Methods using similar forgetting policies described by Yguel et al.~\cite{Yguel2008} cannot adapt to different object classes without compromising the mapping quality and introducing false positives.
Dynablox by Schmid et al.~\cite{Schmid2023} integrates temporal properties in a Truncated Signed Distance Field map, demonstrating generalized performance across diverse dynamic objects. 
Mersch et al.~\cite{Mersch2023} extend 4DMOS with a volumetric approach to retain a memory of spaces that can be occupied by moving objects, increasing the detection rate.
\begin{figure}[t!]
    \centerline{\includegraphics[width=\linewidth]{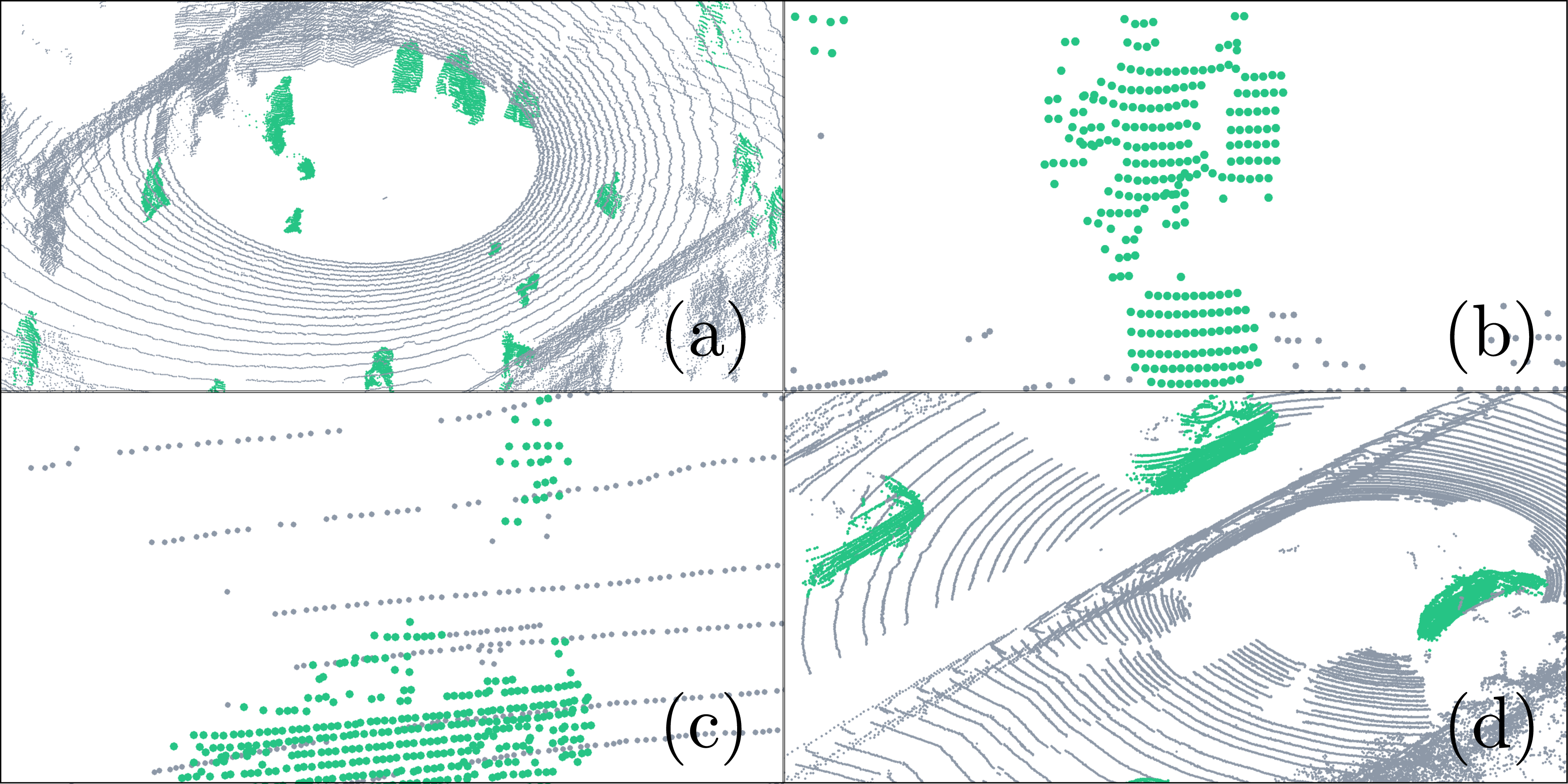}}
    \caption{\small HMM-MOS accurately detects moving objects using the same configuration in all scenarios, including (a) a shopping centre, (b) a person jumping over a moving ball, (c) a pedestrian walking alongside a car, and (d) multiple cars on a highway.}
    \label{fig:hmm_mos_demo}
    \vspace{-6mm}
\end{figure}
\vspace{-3mm}
\section{PROPOSED APPROACH}
\label{sec:proposed_approach}
The general HMM framework applied to identify dynamic objects in point cloud data is adapted from~\cite{James2024}.
This section provides a brief overview of the three stages of the algorithm.

\subsubsection{Voxel Representation}
A map frame, $\mathcal{M}$, is defined to indicate the environment's origin.
At time $k$, the map, $M_{\mathcal{M},k}$, is discretized using voxels, $v$, of a user-configured size $\Delta$.
The voxels are augmented with temporal attributes.
Without uncertainty, detecting dynamic objects is as simple as updating voxel occupancy with new observations, with occupancy changes suggesting dynamic objects.
As the state of each voxel is not directly interpretable due to the associated uncertainty, an HMM is used to represent each voxel's occupancy, similar to~\cite{Meyer-Delius2012, Wang2014}.
Using the notation from~\cite{Rabiner1989}, each voxel is represented using an HMM with three states ($n=3$), $S=\{unobserved, occupied, free\}$.
Let the $i$-th voxel's state vector, $\hat{\mathbf{x}}_{i,k} \in \mathbb{R}^{n\times 1}$, denote the probability of being in each state at time $k$, with the initial state is given by $\hat{\mathbf{x}}_{i,0} = [1,0,0]^\text{T}$.
The state transition matrix, $\mathbf{A}\in \mathbb{R}^{n\times n}$, has large self-transition probabilities for each state, based on the belief that a voxel requires sufficient confidence before transitioning.
The likelihood of the $i$-th voxel being in each state at time $k$ given the sensor observation is encoded in the measurement conditional densities, $\mathbf{B}_{i,k}\in\mathbb{R}^{n\times n}$.
Once defined, a voxel's state is efficiently updated using the recursive HMM filter~\cite{Elliott2008}, $\hat{\mathbf{x}}_{i,k} = \eta_{i,k}\mathbf{B}_{i,k}\mathbf{A}\hat{\mathbf{x}}_{i,k-1}$, where $\eta_{i,k}$ is a normalization that ensures $\hat{\mathbf{x}}_{i,k}$ is a probability.
Voxels outside the sensor's maximum range and those unobserved in the global window, $w_{g}$, are removed from the map.

\subsubsection{Map Update}
\label{sec:new_scan_integration}
A point cloud at time $k$ in the sensor frame, $P_{\mathcal{S},k}$, is transformed by the current sensor pose estimate, $\hat{T}_{\mathcal{M}\rightarrow\mathcal{S},k}$, to locate the scan points in the map frame, $P_{\mathcal{M},k}$.
The scan in map frame, $P_{\mathcal{M},k}$, is discretized at a voxel resolution of $\Delta$, to form a voxelized scan, $P_{\mathcal{M},k}^{'}$.
A raycast is performed using~\cite{Bresenham1965} to find all observed voxels.
All observed voxels are saved in $P_{\mathcal{M},k}^{'obs}$.

The measurement conditional densities of the $i$-th voxel being in a particular state given an observation is, $\mathbf{B}_{i,k} = \text{diag}(0,\mathcal{L}_{v_{i}}^{o},1-\mathcal{L}_{v_{i}}^{o})$, where $\mathcal{L}_{v_{i}}^{o}$,  is the likelihood of the voxel being occupied.
An observed voxel is likely to be occupied if it is close to a voxel in the voxelized scan, $P_{\mathcal{M},k}^{'}$, and free otherwise.
This is captured using the scan's Euclidean Distance Field~\cite{Oleynikova2016}.
The EDF value for the $i$-th observed voxel, $d_{i}$, is used to calculate the occupancy likelihood by evaluating an unnormalized Gaussian at $d_i$, $\mathcal{L}_{v_{i}}^{o} = \exp{(-d_{i}^2/2\sigma_{o}^2)}$, where $\sigma_{o}$ is a user-configured standard deviation to capture uncertainty in the estimate.
A voxel's state is updated when the state's probability surpasses a predefined threshold, $p_{min}$.

\subsubsection{Dynamic Point Identification}
A voxel's occupancy transition seeds the detection of dynamic objects.
The first step is to identify voxels from the current voxelized scan, $P_{\mathcal{M},k}^{'}$, that changed state in the voxel map, $M_{\mathcal{M},k}$, captured in $P_{\mathcal{M},k}^{'chg}$.
The change detection allows for likely dynamic voxels to be identified, however, changes in the voxel's neighbourhood are not examined.
A spatiotemproal (4D) convolution is performed to identify missed detections and suppress noisy detections.
For each voxel, $v_{i}\in P_{\mathcal{M},k}^{'}$, the likelihood of being dynamic, $\mathcal{L}_{v_{i}}^{dyn}$, is calculated by summing state changes in the voxel's local neighbourhood over a local window size of $w_{l}$.
A kernel, $K_{m}\in\mathbb{R}^{m\times m\times m}$, is convolved with each voxel in $P_{\mathcal{M},(k-w_{l})\rightarrow k}^{'}$ to compute $\mathcal{L}_{v_{i}}^{dyn}$.

A voxel's dynamic likelihood depends on the voxel size, the convolution kernel, the scan sparsity, and the object.
Hence, manually thresholding to extract dynamic voxels based on their likelihood is challenging.
Otsu's automatic thresholding~\cite{Otsu1979} is applied to the convolution scores to extract the set of dynamic voxels, $P_{\mathcal{M},k}^{'dyn}$.
High-confidence dynamic voxels from the previous scan are preserved in the current scan and saved in a temporal dynamic occupancy map of size $w_{d}$ scans to assist with future detections.
A nearest neighbour dilation is applied to $P_{\mathcal{M},k}^{'dyn}$ to grow the dynamic detection results into neighbouring regions.

\section{RESULTS}
\label{sec:results}
We evaluate the proposed algorithm using the DOALS~\cite{Pfreundschuh2021}, Sipailou Campus~\cite{Zhou2023}, and HeLiMOS~\cite{Lim2024} benchmark datasets to test generalized performance.
All tests use the same configuration with an uncertainty equal to the voxel size, $\sigma_{o}=\Delta$, $p_{min}=0.99$, a convolution kernel size of $m=5$, $otsu_{min}=3$, and scan windows of $w_{l}=3$, $w_{d}=100$, and $w_{g}=300$ to demonstrate generalized behaviour.
The full testing conditions, sensor poses, sample videos, and all results are linked on our open-source page.
All tests estimate sensor pose using~\cite{BhandariSimple}, unless stated otherwise.

The results displayed in Tables I-III demonstrate the strong generalization capabilities of the proposed algorithm in comparison to state-of-the-art methods.
The DOALS dataset is recorded with a handheld OS1-64 in environments with diverse dynamic objects predominantly consisting of pedestrians, the Sipailou Campus dataset is recorded using a Livox Avia mounted to an unmanned ground vehicle as it traverses a university campus, whereas the new HeLiMOS dataset is recorded with four different LiDARs mounted to a vehicle in dynamic urban environments.
We demonstrate consistent performance, on par with or performing better than state-of-the-art such as Dynablox~\cite{Schmid2023}, 4DMOS~\cite{Mersch2022} and MapMOS~\cite{Mersch2023}.
To evaluate generalized performance, we compare the HeLiMOS benchmark results with the methods trained on the SemanticKitti dataset.
When the benchmark approaches are trained on the new data, they outperform the proposed approach, see~\cite{Lim2024}.
The algorithm's performance metrics are hindered by the ground truth labelling process as we detect movement only, and not if the object has moved throughout the scan sequence.
This severely decreases the recall.
The proposed algorithm is computationally expensive and only provides real-time results within a 20-50m range depending on the point cloud density.
There is ongoing work to achieve real-time results for larger detection ranges.
\vspace{-3mm}
\begin{table}[h]
    \centering
    \caption{\small Evaluation on the DOALS dataset with best results in bold. Results for other methods are as documented by~\cite{Schmid2023}.}
    \begin{tabular}{lcccc}
        \hline
        \textbf{Method} & \textbf{ST} & \textbf{SV} & \textbf{HG} & \textbf{ND}\\
        \hline
        4DMOS~\cite{Mersch2022} & 38.8 & 50.6 & 71.1 & 40.2\\
        LMNet~\cite{Chen2021} (Refit) & 19.9 & 18.9 & 27.4 & 40.1\\
        Dynablox~\cite{Schmid2023} & \textbf{86.2} & \textbf{83.2} & 84.1 & \textbf{81.6} \\
        This paper (online), $\Delta=0.20m$ & 82.7 & 80.8 & \textbf{85.9} & 81.4\\
        \hline
        LC Free Space~\cite{Modayil2008} (20\,m) & 48.7 & 31.9 & 24.7 & 17.7\\
        Dynablox~\cite{Schmid2023} (20\,m) & 87.3 & \textbf{87.8} & 86.0 & 83.1\\
        This paper (online), $\Delta=0.20m$ (20\,m) & \textbf{88.9} & 84.7 & \textbf{87.3} & \textbf{83.5}\\
        \hline
    \end{tabular}
    \label{tab:results_doals}
\end{table}
\vspace{-5mm}
\begin{table}[h]
    \centering
    \caption{\small Evaluation on the Sipailou Campus dataset with best results in bold. Results for other methods are as reported by~\cite{Zhou2023}.}
    \begin{tabular}{lcc}
        \hline
        \textbf{Method} & \textbf{IoU Validation} & \textbf{IoU Test} \\
        \hline
        MotionSeg3D~\cite{Sun2022} & 6.83 & 6.72 \\
        4DMOS~\cite{Mersch2022} & 78.54 & 82.30 \\
        Motion-BEV-h~\cite{Zhou2023} & 70.94 & 71.51 \\
        This paper (online), $\Delta=0.25m$ & \textbf{85.60} & \textbf{87.00} \\ 
        \hline
    \end{tabular}
    \label{tab:results_Sipailou_campus}
\end{table}
\vspace{-5mm}
\begin{table}[h]
    \centering
    \caption{\small Evaluation on the HeLiMOS dataset with best results in bold. Results for other methods are as documented by~\cite{Lim2024}.}
    \begin{tabular}{lccccc}
        \hline
        \textbf{Method} & \textbf{L} & \textbf{A} & \textbf{O} & \textbf{V} & \textbf{Avg}\\
        \hline
        4DMOS, online~\cite{Mersch2022} & 52.1 & 54.0 & 64.2 & 4.7 & 43.7\\
        4DMOS, delayed~\cite{Mersch2022} & 59.0 & 58.3 & 70.4 & 5.4 & 48.3\\
        MapMOS, Scan~\cite{Mersch2023} & 58.9 & 63.2 & 81.4 & 4.3 & 52.0\\
        MapMOS, Volume~\cite{Mersch2023} & \textbf{62.7} & 66.6 & \textbf{82.9} & 5.8 & 54.5\\
        This paper (online), $\Delta=0.25m$ & 51.3 & 69.8 & 75.0 & 35.0 & 57.8\\
        This paper, delayed, $\Delta=0.25m$ & 57.6 & \textbf{70.0} & 73.4 & \textbf{53.9} & \textbf{63.7}\\
        \hline
    \end{tabular}
    \label{tab:results_Helimos}
\end{table}

\vspace{-3mm}
\section{CONCLUSIONS}
\label{sec:conclusions}
This paper presents a learning-free solution to the MOS problem.
The significance of the work is that it is robust and generalizes to a range of datasets without reconfiguration, producing comparable or better results than state-of-the-art.

\bibliography{ref.bib}
\bibliographystyle{IEEEtran}

\end{document}